\documentclass{article}
\usepackage{iclr2025_conference,times}

\usepackage{amsmath,amsfonts,bm}

\def\eqref#1{equation~\ref{#1}}

\def\1{\bm{1}}

\DeclareMathAlphabet{\mathsfit}{\encodingdefault}{\sfdefault}{m}{sl}
\SetMathAlphabet{\mathsfit}{bold}{\encodingdefault}{\sfdefault}{bx}{n}

\usepackage{url}
\usepackage{enumitem}
\usepackage{graphicx}
\usepackage{tikz}
\usetikzlibrary{arrows.meta,positioning,calc,fit,matrix}
\usepackage{xcolor}
\definecolor{FixedRed}{HTML}{8E1E2E} 
\usepackage{hyperref}
\usepackage[nameinlink,noabbrev,capitalise]{cleveref}

\iclrfinalcopy

\title{Decoupling Search and Learning in Neural Net Training}
\author{
\begin{minipage}[t]{0.45\textwidth}\centering
{\bfseries Akshay Vegesna}\thanks{Equal contribution. Correspondence to \texttt{research@qlabs.sh}}\\
{\normalfont Q Labs}\\
\end{minipage}
\And
\begin{minipage}[t]{0.45\textwidth}\centering
{\bfseries Samip Dahal}\footnotemark[1]\\
{\normalfont Q Labs}\\
\end{minipage}
}

\begin{document}
\pagestyle{plain}  
\maketitle
\thispagestyle{plain}

\begin{abstract}
Gradient descent typically converges to a single minimum of the training loss without mechanisms to explore alternative minima that may generalize better. Searching for diverse minima directly in high-dimensional parameter space is generally intractable. To address this, we propose a framework that performs training in two distinct phases: search in a tractable representation space (the space of intermediate activations) to find diverse representational solutions, and gradient-based learning in parameter space by regressing to those searched representations. Through evolutionary search, we discover representational solutions whose fitness and diversity scale with compute—larger populations and more generations produce better and more varied solutions. These representations prove to be learnable: networks trained by regressing to searched representations approach SGD's performance on MNIST, CIFAR-10, and CIFAR-100. Performance improves with search compute up to saturation. The resulting models differ qualitatively from networks trained with gradient descent, following different representational trajectories during training. This work demonstrates how future training algorithms could overcome gradient descent's exploratory limitations by decoupling search in representation space from efficient gradient-based learning in parameter space.
\end{abstract}

\section{Introduction}
Neural network training is fundamentally a search process over parameter configurations, seeking those that minimize training loss while generalizing well to unseen data. The ideal approach for generalization would be exhaustive search—systematically exploring the parameter space to discover many diverse minima, then selecting those with the best generalization properties. However, the parameter space of modern neural networks is vast, containing millions or billions of dimensions. Exhaustive search of this space is computationally intractable; even with substantial compute, we could only explore a vanishingly small fraction of possible configurations.
Gradient descent emerged as a practical alternative to this intractable search problem. Rather than exploring broadly, it efficiently descends from an initial point by following local gradients, finding good solutions without needing to examine the entire parameter space. This efficiency comes with a fundamental limitation: gradient descent converges to a single local minimum and cannot explore alternative regions of the loss landscape to discover the diverse solutions that may exist elsewhere. This trade-off between exhaustive search (diverse but intractable) and gradient descent (efficient but limited) motivates the need for training algorithms that can combine both approaches.

We argue that exploring diverse solutions may be fundamental to achieving better generalization. To understand why, consider what a learning algorithm that yields optimal generalization would do. Our core hypothesis (detailed \href{https://getcue.ai/research}{here}) is that given infinite compute, such an algorithm would enumerate all models consistent with the data and use a generalization prior to select the best one. Solomonoff induction \citep{solomonoff1964formal} is an example of this ideal—enumerating all computable models and weighting them by simplicity. While computationally intractable, this framework reveals what's missing from gradient-based training: exploration of diverse solutions through search.

Though complete synthesis of search and gradient descent extends beyond a single paper, we take a first step by decomposing the problem: instead of searching directly in parameter space, we perform explicit search in representation space—the activations of intermediate layers. This space is far smaller, making search tractable via evolutionary algorithms. Moreover, these searched representations prove to be learnable: neural networks can be trained through gradient descent to predict these representations.
This means we can find effective intermediate representations through search, then train network parameters to produce those patterns—\emph{effectively using search to guide where gradient descent should go}. By scaling search to discover diverse representational solutions and steering gradient descent toward them, future algorithms could overcome gradient descent's inability to explore.

This paper takes three steps:
\begin{enumerate}[leftmargin=1.5em]
\item \textbf{Conceptual framework} (\cref{sec:framework}): We recast neural network training as two decoupled phases. First, we perform evolutionary search in the representation space to discover representational solutions. Second, we use gradient-based learning in parameter space to train the network to produce these representations. This separation preserves the benefits of search while avoiding the computational intractability of searching in parameter space.

\item \textbf{Search in representation space} (\cref{sec:search}): We implement evolutionary search over representations to minimize training loss and study how it scales with compute. We find that both solution fitness and diversity increase with more compute—more generations and larger populations yield better and more varied representational solutions.

\item \textbf{Learning in parameter space} (\cref{sec:learning}): We train network parameters to match the searched representations and show this method approaches the performance of SGD on MNIST, CIFAR-10, and CIFAR-100—all without backpropagating cross-entropy gradients through the network body. We analyze how performance scales with search compute and demonstrate that the resulting models are qualitatively different from those produced by SGD.
\end{enumerate}

\section{Conceptual Framework}
\label{sec:framework}

The most direct way to instantiate our hypothesis—searching over many models and selecting good ones based on the generalization prior—would be to search directly in parameter space. However, the choice of search space must satisfy specific principles for both search and learning.

\subsection{Core principles for search and learning}

For effective search, we argue that the space must satisfy two key principles:
\begin{enumerate}[leftmargin=1.5em]
\item Random search tractability. The space must be small enough that naive random sampling or perturbations can make progress towards high-fitness solutions.
\item Amortization. We want to train a sampler that learns from random perturbations and their fitness values, generalizing to produce new high-fitness solutions. We can then search around these solutions with further perturbations and amortize again, creating an iterative improvement cycle.
\end{enumerate}

When a space doesn't satisfy these search principles, we fall back to gradient-based optimization when applicable.

\subsection{Why parameter space fails for search}

Parameter space fails both search requirements. First, it is far too large for random sampling to be effective—the probability of randomly finding good parameters is vanishingly small. Second, learning to directly sample parameters of a large neural net is difficult, although this has been tried at a small scale \citep{wang2024neural}.

Given these limitations, parameter space only supports learning through gradient descent, not search.

\subsection{Decoupling}
Good models must produce good representations. This observation suggests we can decompose the problem: instead of searching for parameters directly, we can search for good representations, then learn parameters that produce those representations using gradient descent. This breaks the intractable problem of searching in parameter space into two tractable subproblems.

Representation space satisfies both search principles. First, the space of activations is far smaller than parameter space, making random perturbations effective at finding high-fitness solutions. Second, neural networks already define predictive distributions over their own representations, providing a built-in mechanism for amortization.

Once we have discovered high-fitness representations through search, learning parameters to induce them is straightforward with gradient descent.

\section{Search}
\label{sec:search}
\subsection{How we search in representation space}
\label{subsec:how-search}

\paragraph{Core algorithm.}
We perform evolutionary search directly on neural network representations by searching over the layerwise activations—the actual intermediate tensors produced at key layers. Rather than optimizing parameters, we treat the activation tensors at selected layers as the primary search space, evolving these representations to minimize classification loss. The key insight is that by evolving representations at intermediate layers in a forward pass, we can discover high-quality solutions without backpropagation.

Our approach consists of two main stages: (1) initialize a population by forward propagating inputs with noise, and (2) sequentially evolve representations at each selected layer while fixing earlier ones. We evolve activations for the first convolutional block output, fix them, then evolve the next block's activations that build on these fixed representations, and so on through the network. The optimized activations found through search then serve as regression targets to train the network parameters.

\paragraph{Network architecture.}
We apply our method to a standard convolutional network for CIFAR-10 classification. The network consists of three convolutional blocks followed by a linear classification head. Each block contains two \(3 \times 3\) convolutional layers with batch normalization and GELU activations, with \(2 \times 2\) max pooling between blocks to progressively reduce spatial dimensions. The final convolutional features are flattened and fed to a linear layer that outputs class logits. The three blocks produce feature maps of dimensions \(256 \times 15 \times 15\), \(256 \times 7 \times 7\), and \(256 \times 3 \times 3\) respectively, giving us four representation levels to optimize (three convolutional outputs and final logits). We perform search using a network initialized with Dirac initialization for convolutional layers and Kaiming Uniform initialization for other layers.

\paragraph{Problem setup.}
Given input batch $(x, y)$ and a network, we perform evolutionary search at $L=4$ specific points in the architecture: the outputs of the three convolutional blocks ($\ell \in \{0, 1, 2\}$) and the final logits ($\ell = 3$). We refer to these architectural points as search layers since each convolutional block contains multiple internal layers but we only perform search at the block outputs. For the remainder of this section and the next, layer refers to these search layers unless otherwise specified. Let $H^{(\ell)}$ denote the activations at layer $\ell$. The population at layer $\ell$ is $\mathcal{P}_\ell = \{H_i^{(\ell)}\}_{i=1}^{n_{\text{pop}}}$, representing $n_{\text{pop}}$ candidate activations. We use negative cross-entropy as our fitness function (equivalently, maximizing log-likelihood), evaluating how well each candidate's final predictions match the true labels. All selection, mutation, and fitness evaluation operations are performed independently for each image in the batch.

\paragraph{Population initialization.}
We initialize the population at layer 0 by sampling \(n_{\text{pop}}\) noisy variants of the inputs to the first layer. The noise is channel-wise and scaled to input statistics, yielding \(n_{\text{pop}}\) distinct \(H^{(0)}\) samples that serve as the starting point for evolution.

\begin{figure}[t]
\centering

\begin{minipage}{0.32\linewidth}
\centering
\begin{tikzpicture}[font=\small,>=Latex]
\tikzset{
  blk/.style={draw, rounded corners, thick, minimum width=2.8cm, minimum height=0.8cm, align=center, fill=white},
  rep/.style={draw, circle, thick, minimum size=7mm, align=center, inner sep=1pt, fill=white},
  repmini/.style={draw, circle, thick, minimum size=4.2mm, align=center, inner sep=0.5pt, fill=white},
  popbox/.style={draw, dotted, rounded corners, thick, inner sep=5pt}
}

\node[blk] (inL)                {Input $x$};
\node[blk, below=0.6cm of inL]  (b0L) {Conv Block 0};

\node (manchorL) [below=0.65cm of b0L] {};
\matrix (mL) [matrix of nodes,
             nodes={repmini},
             column sep=1mm,
             anchor=north] at (manchorL)
{
  $H^{(0)}_{1}$ & $H^{(0)}_{2}$ & $H^{(0)}_{n}$ \\
};
\node[popbox, fit=(mL-1-1)(mL-1-3),
      label={[font=\scriptsize,anchor=south west,xshift=2pt]north west:$\mathcal{P}_0$}] (pop0L) {};

\node[blk, below=0.75cm of pop0L] (b1L) {Conv Block 1};

\draw[->, thick] (inL) -- (b0L);
\draw[->, thick] (b0L) -- (pop0L.north);
\draw[->, thick] (pop0L.south) -- (b1L);
\end{tikzpicture}

\vspace{0.35em}
\small (a) Evolve population $\mathcal{P}_0$.
\end{minipage}
\hfill
\begin{minipage}{0.32\linewidth}
\centering
\begin{tikzpicture}[font=\small,>=Latex]
\tikzset{
  blk/.style={draw, rounded corners, thick, minimum width=2.8cm, minimum height=0.8cm, align=center, fill=white},
  rep/.style={draw, circle, thick, minimum size=7mm, align=center, inner sep=1pt, fill=white},
  repmini/.style={draw, circle, thick, minimum size=4.2mm, align=center, inner sep=0.5pt, fill=white},
  repfixed/.style={
      draw,
      circle,
      thick,
      minimum size=7mm,
      align=center,
      inner sep=1pt,
      fill=FixedRed,
      text=white
  },
  popbox/.style={draw, dotted, rounded corners, thick, inner sep=5pt}
}

\node[blk] (inR)                {Input $x$};
\node[blk, below=0.6cm of inR]  (b0R) {Conv Block 0};
\node[repfixed, below=0.55cm of b0R] (h0R) {$H^{(0)}$};
\node[blk, below=0.6cm of h0R]  (b1R) {Conv Block 1};

\node (manchorR) [below=0.65cm of b1R] {};
\matrix (mR) [matrix of nodes,
             nodes={repmini},
             column sep=1mm,
             anchor=north] at (manchorR)
{
  $H^{(1)}_{1}$ & $H^{(1)}_{2}$ & $H^{(1)}_{n}$ \\
};
\node[popbox, fit=(mR-1-1)(mR-1-3),
      label={[font=\scriptsize,anchor=south west,xshift=2pt]north west:$\mathcal{P}_1$}] (pop1R) {};

\node[blk, below=0.75cm of pop1R] (b2R) {Conv Block 2};

\draw[->, thick] (inR) -- (b0R);
\draw[->, thick] (b0R) -- (h0R);
\draw[->, thick] (h0R) -- (b1R);
\draw[->, thick] (b1R) -- (pop1R.north);
\draw[->, thick] (pop1R.south) -- (b2R);
\end{tikzpicture}

\vspace{0.35em}
\small (b) Evolve population $\mathcal{P}_1$ with $H^{(0)}$ fixed.
\end{minipage}
\hfill
\begin{minipage}{0.32\linewidth}
\centering
\begin{tikzpicture}[font=\small,>=Latex]
\tikzset{
  blk/.style={draw, rounded corners, thick, minimum width=2.8cm, minimum height=0.8cm, align=center, fill=white},
  rep/.style={draw, circle, thick, minimum size=7mm, align=center, inner sep=1pt, fill=white},
  repmini/.style={draw, circle, thick, minimum size=4.2mm, align=center, inner sep=0.5pt, fill=white},
  repfixed/.style={
      draw,
      circle,
      thick,
      minimum size=7mm,
      align=center,
      inner sep=1pt,
      fill=FixedRed,
      text=white
  },
  popbox/.style={draw, dotted, rounded corners, thick, inner sep=5pt}
}

\node[blk] (inRR)                {Input $x$};
\node[blk, below=0.6cm of inRR]  (b0RR) {Conv Block 0};
\node[repfixed, below=0.55cm of b0RR] (h0RR) {$H^{(0)}$};
\node[blk, below=0.6cm of h0RR]  (b1RR) {Conv Block 1};
\node[repfixed, below=0.55cm of b1RR] (h1RR) {$H^{(1)}$};
\node[blk, below=0.6cm of h1RR]  (b2RR) {Conv Block 2};

\node (manchorRR) [below=0.65cm of b2RR] {};
\matrix (mRR) [matrix of nodes,
              nodes={repmini},
              column sep=1mm,
              anchor=north] at (manchorRR)
{
  $H^{(2)}_{1}$ & $H^{(2)}_{2}$ & $H^{(2)}_{n}$ \\
};
\node[popbox, fit=(mRR-1-1)(mRR-1-3),
      label={[font=\scriptsize,anchor=south west,xshift=2pt]north west:$\mathcal{P}_2$}] (pop2RR) {};

\draw[->, thick] (inRR) -- (b0RR);
\draw[->, thick] (b0RR) -- (h0RR);
\draw[->, thick] (h0RR) -- (b1RR);
\draw[->, thick] (b1RR) -- (h1RR);
\draw[->, thick] (h1RR) -- (b2RR);
\draw[->, thick] (b2RR) -- (pop2RR.north);

\end{tikzpicture}

\vspace{0.35em}
\small (c) Evolve population $\mathcal{P}_2$ with $H^{(0)}$, $H^{(1)}$ fixed.
\end{minipage}

\caption{Layerwise forward evolution for $H^{(0)}$, $H^{(1)}$, and $H^{(2)}$. 
At each layer $\ell$, we evolve a population $\mathcal{P}_\ell$ (shown within dashed boxes), keep the best as $H^{(\ell)}$ (red), and fix it for subsequent layers.
After mutation, fitness is evaluated by completing the remaining forward pass. 
Representation sizes: $H^{(0)}\!\in\!\mathbb{R}^{256\times15\times15}$, $H^{(1)}\!\in\!\mathbb{R}^{256\times7\times7}$, $H^{(2)}\!\in\!\mathbb{R}^{256\times3\times3}$.}
\label{fig:arch-compact-layer012}
\end{figure}
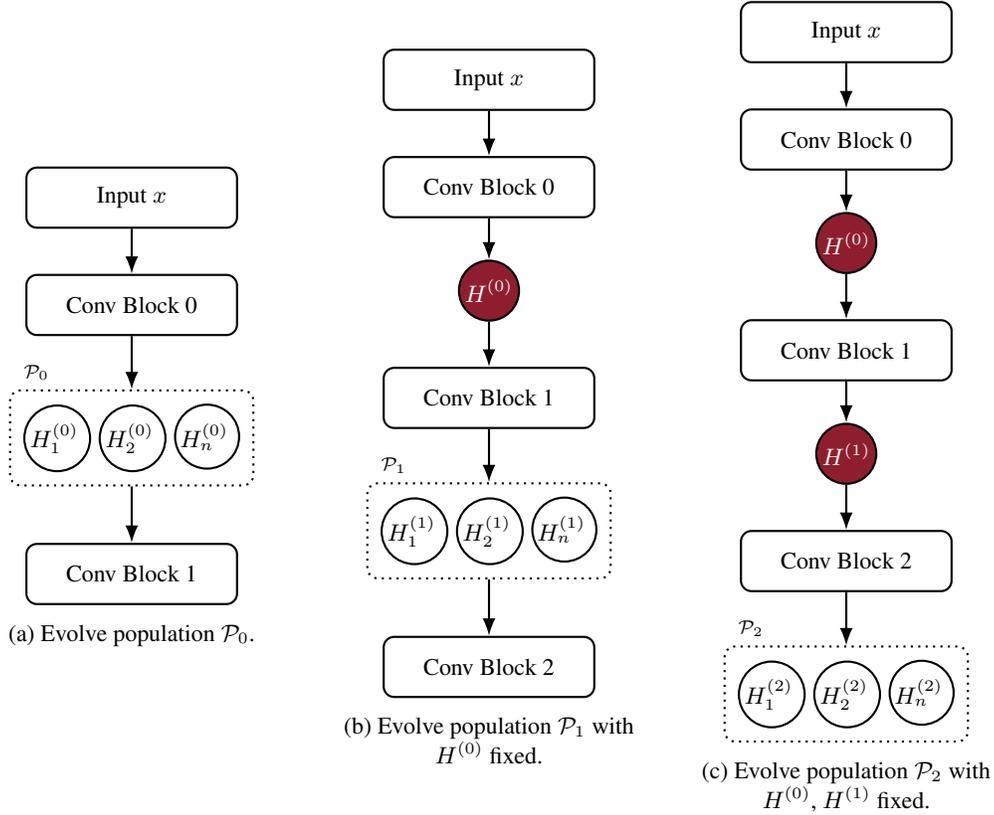

\paragraph{Layer-wise forward evolution.}
We evolve layers sequentially from \(\ell = 0\) to \(L-1\). When evolving layer \(\ell\), we fix the representations at all earlier layers \(H^{(0)}, \ldots, H^{(\ell-1)}\) to their previously evolved values, evolve only \(H^{(\ell)}\), then forward propagate through the remaining network layers to recompute representations at later layers. This ensures that each layer's search builds upon the optimized solutions found for previous layers, creating a compositional optimization process where later layers benefit from earlier improvements. \cref{fig:arch-compact-layer012} illustrates this layer-wise evolution process, showing how we progressively evolve populations at each layer while fixing earlier layer representations (shown in red).

\paragraph{Evolution mechanics.}
For each generation at layer \(\ell\), we select the top-\(k\) candidates from the population independently for each image in the batch based on fitness. This per-image selection (rather than batch-wide) accelerates convergence by maintaining diversity across different data points. From these parents, we: (1) retain all \(k\) parents unchanged, (2) create \(C_{\text{exp}}\) exploratory samples with high mutation strength (\(\alpha \times\) exploration boost) to discover new high-fitness regions and avoid diversity collapse, and (3) generate \(C_{\text{ref}}\) refinement samples with standard mutation strength (\(\alpha\)) for local improvement. The next generation consists of parents \(\cup\) exploratory samples \(\cup\) refinement samples.

\paragraph{Reproduction operators.}  
The exploratory and refinement samples are created using the same three-step process: genetic modification at layer~$\ell$, forward propagation, and fitness evaluation. The only difference is the mutation strength~$\alpha$.  

For genetic modification, we first apply crossover $\tilde{H}^{(\ell)} = \tfrac{1}{2}(H_{p_1}^{(\ell)} + H_{p_2}^{(\ell)})$, where $p_1, p_2$ are randomly selected parents. We then add Gaussian mutation $\epsilon \sim \mathcal{N}(0, (\alpha \sigma_{\text{parent}})^2)$, where $\sigma_{\text{parent}}$ is the per-channel standard deviation of the parent activations. For convolutional layers, we use channel-wise noise (preserving spatial coherence), apply spatial smoothing via repeated $3 \times 3$ average pooling (to reduce high-frequency artifacts), and normalize to zero mean and unit variance. We found spatial smoothing to be crucial for learnability of these representations with gradient descent, and normalization for convergence of evolution. For logits, we simply add element-wise noise and recenter. 

After modifying layer \(\ell\), we forward propagate through the network's blocks to compute \(H^{(\ell+1)},\ldots,H^{(L-1)}\), then evaluate fitness using the final classification loss. This ensures every mutation is evaluated in the context of the full network.

\paragraph{Pseudocode (Python-style)}

\begin{verbatim}
# Inputs: x, y, pop_size, top_k, c_exp, c_ref, gens_per_layer

# Initialize population with channel-wise noise
population = []
for _ in range(pop_size):
    h0 = x + channel_noise(x)
    population.append(propagate_through_network(h0))
fitness = evaluate_fitness(population, y)

# Evolve each layer sequentially
for layer in range(num_layers):
    for gen in range(gens_per_layer[layer]):
        # Select top-k parents per image
        parents = select_top_k_per_image(population, fitness, k=top_k)
        
        # Exploration: high mutation strength (alpha * boost)
        explorers = reproduce(parents, n=c_exp, mutation="high")
        
        # Refinement: standard mutation strength (alpha)
        refiners = reproduce(parents + explorers, n=c_ref, mutation="standard")
        
        # New population with downstream layers recomputed
        population = parents + explorers + refiners
        population = recompute_from_layer(population, start=layer+1)
        fitness = evaluate_fitness(population, y)
    
    # Keep best individual for next layer
    population = keep_best(population)
\end{verbatim}

\subsection{Scaling with Compute}
\label{subsec:compute-scaling}
\paragraph{Fitness scaling.}
We study how search quality improves with compute by varying either the population size or the number of generations while holding the other fixed. 
For each convolutional block representation $\ell\!\in\!\{0,1,2\}$ we evolve only $H^{(\ell)}$, then recompute $H^{(\ell+1)},\ldots,H^{(L-1)}$ before scoring fitness (we report cross-entropy; lower is better). 
For each configuration we average the best individual’s loss per image across $1000$ random data points in the training set of CIFAR-100. 

\begin{figure}[t]
\centering
\begin{minipage}{0.49\linewidth}
    \includegraphics[width=\linewidth]{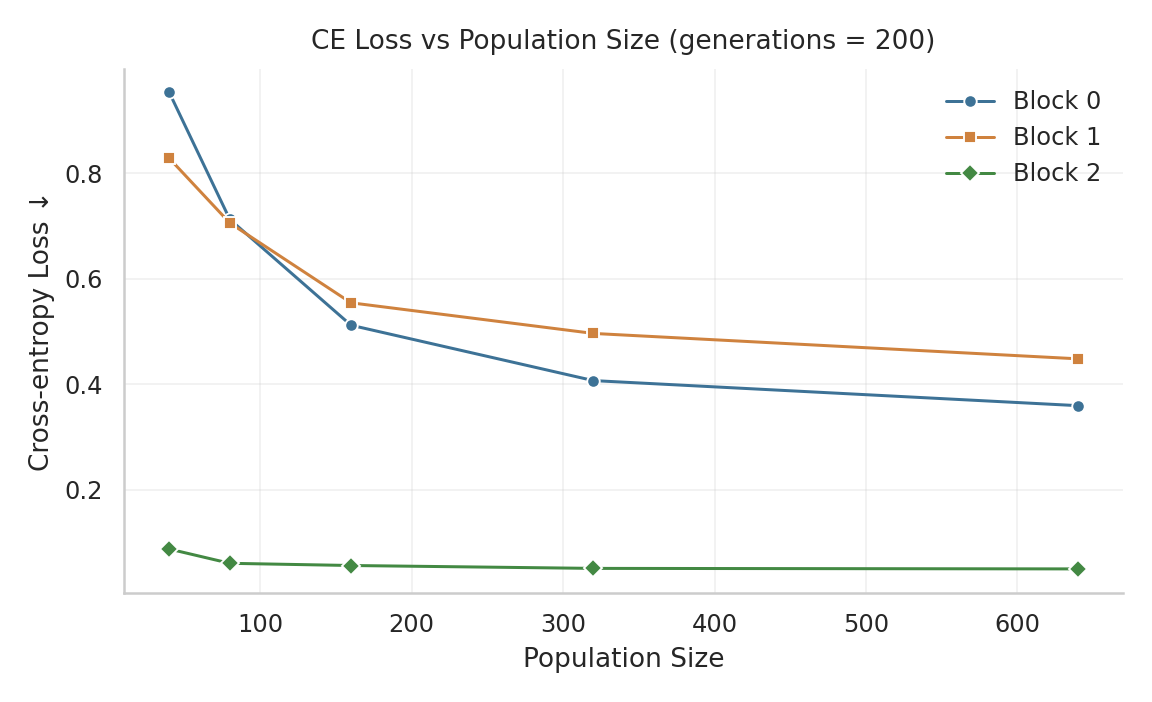}
\end{minipage}\hfill
\begin{minipage}{0.49\linewidth}
    \includegraphics[width=\linewidth]{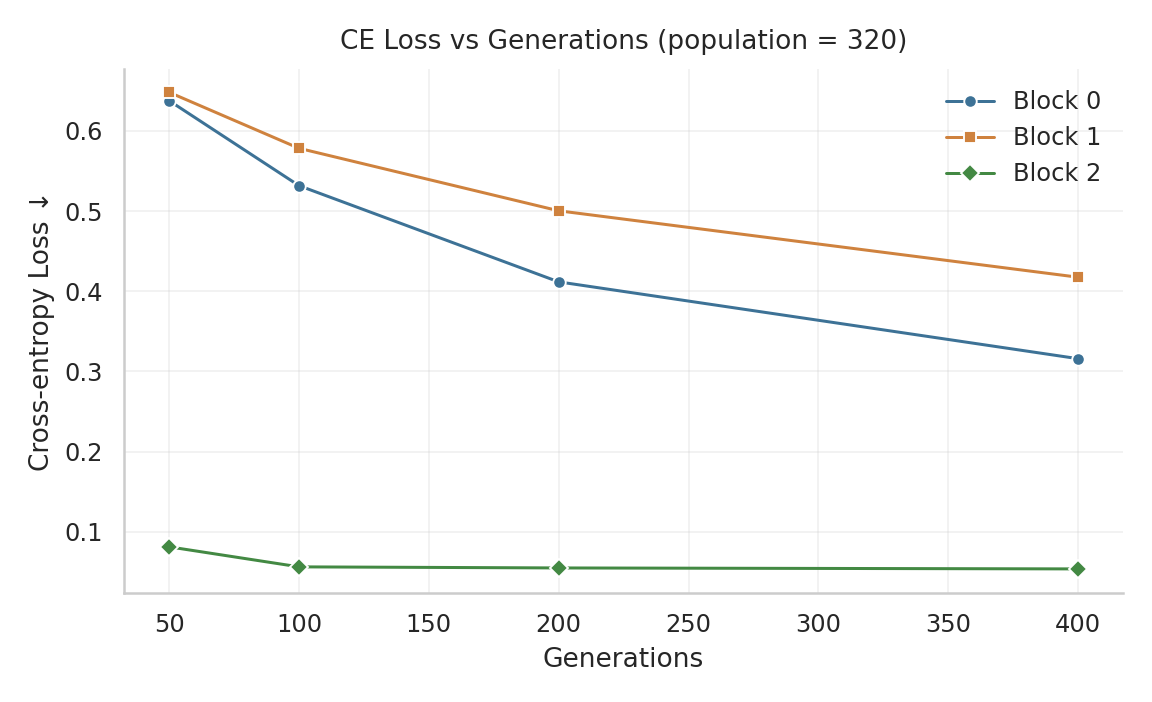}
\end{minipage}
\caption{Fitness scales with compute on CIFAR-100.
Mean cross-entropy vs.\ population size (left) and vs.\ generations (right).}
\label{fig:search-scaling}
\end{figure}

\cref{fig:search-scaling} shows that loss decreases with more compute along both axes—population size and number of generations. 
As expected, the later layer (Block 2) optimizes more easily than the earlier ones (Blocks 0 and 1).

\paragraph{Diversity scaling.} 
We also study how the diversity of solutions produced by evolution scales with compute. For this, we perform multiple independent evolutionary search runs and collect the top candidate from each run, provided its predicted probability for the correct class exceeds 0.5. Following prior work \citep{skean2025layer}, we then compute the effective number of distinct solutions, $N_{\text{eff}}$, derived from the collision entropy of the cosine similarity Gram matrix:
\[
H_{\mathrm{coll}} = -\log \frac{\sum_{i,j} K_{ij}^2}{\left(\sum_i K_{ii}\right)^2}, 
\qquad N_{\text{eff}} = \exp(H_{\mathrm{coll}}),
\]
where $K$ is the pairwise cosine similarity matrix of candidate representations. Intuitively, $N_{\text{eff}}$ is the effective number of completely orthogonal solutions; in practice, there usually are many more partially overlapping solutions, so this quantity is a conservative estimate of diversity.

\begin{figure}[t]
\centering
\includegraphics[width=0.7\linewidth]{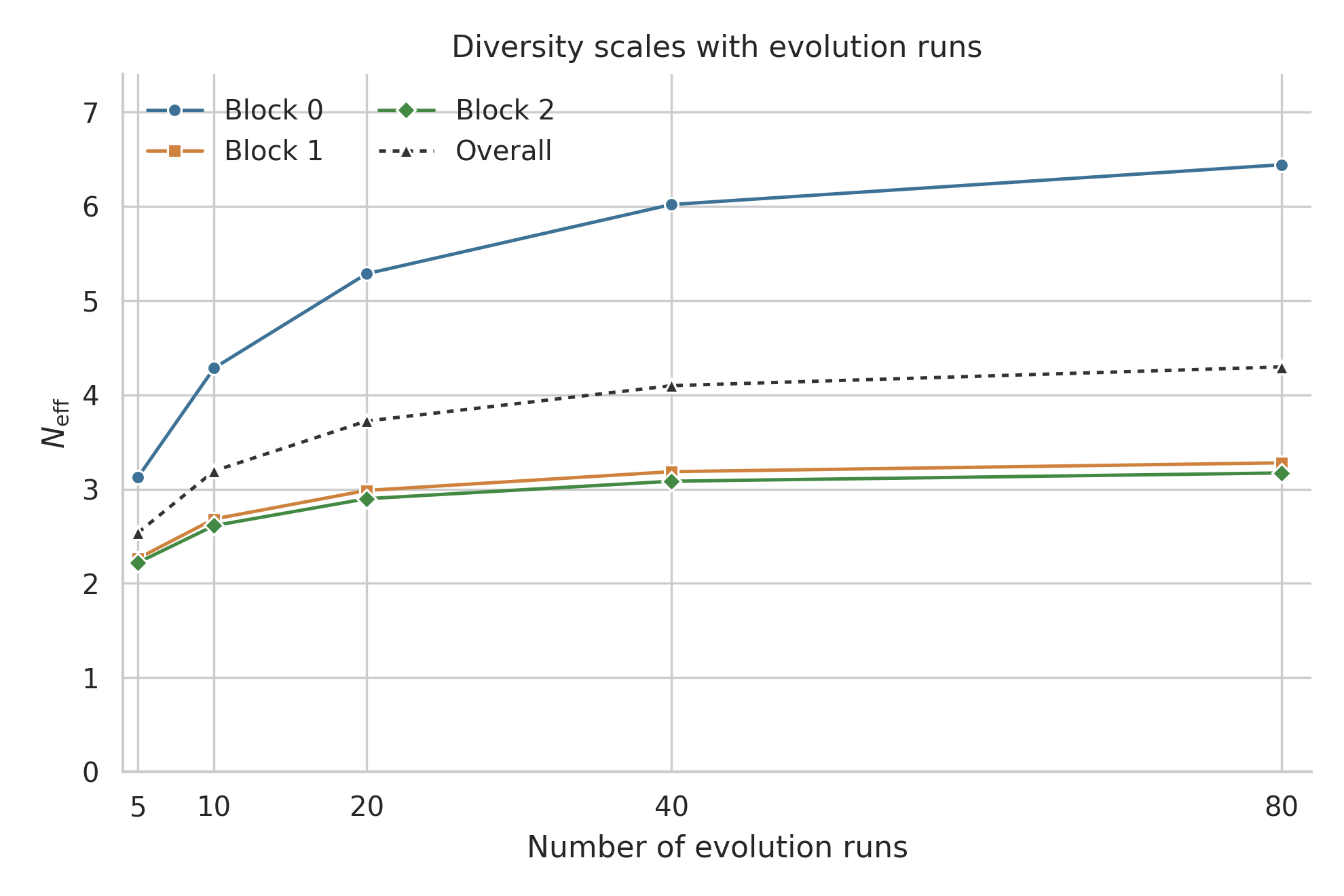}
\caption{Diversity grows with compute across convolutional block representations $\ell \in \{0, 1, 2\}$ on CIFAR-100. Effective number of solutions $N_{\text{eff}}$ as a function of the number of independent evolutionary runs.}
\label{fig:diversity-scaling}
\end{figure}

\cref{fig:diversity-scaling} shows that $N_{\text{eff}}$ steadily increases as we aggregate results from more independent evolutionary search runs. Early convolutional block representations show the strongest growth in $N_{\text{eff}}$, consistent with their larger representational capacity, while later blocks exhibit slower growth. 

This demonstrates that search naturally produces both optimization and diversity scaling with compute.

\section{Learning}
\label{sec:learning}

\subsection{Why representations alone are not enough}

Searching representations by itself is not sufficient. Two further requirements must hold:
(i) the searched representations must be learnable by the network’s layers, and
(ii) learning those representations must lead to generalization on unseen data. In other words, searched representations should provide a path to models that generalize. Our hypothesis is that regressing to searched representations yields such models, and regressing to different representational solutions yields different model solutions.

\subsection{Learning from searched representations}
\label{sec:learning-setup}

\paragraph{Caching.}
We run evolutionary search once over the training set and cache the results. The search is performed using an untrained initialized network. For each training example $(x,y)$, we store the best solution from search: the full sequence of representations $\{\hat{H}^{(\ell)}\}_{\ell=0}^{L-1}$, with the searched output probability distribution defined from the searched logits by $\hat{p}(\cdot \mid x) = \mathrm{softmax}(\hat{H}^{(L-1)}(x))$. These cached representations become fixed regression targets—we never re-run search during training.

\paragraph{Objective.}
We design our training objective to ensure the convolutional body learns exclusively from the searched representations, not from classification gradients. We minimize the following with gradient descent:
\[
\mathcal{L}(\theta,\phi) \;=\;
\frac{1}{L-1}\sum_{\ell=0}^{L-2}
\frac{1}{B}\sum_{b=1}^B
\big\| H^{(\ell)}_\theta(x_b) - \hat{H}^{(\ell)}(x_b) \big\|_2^2
\;+\;
\frac{\lambda}{B}\sum_{b=1}^B
\mathrm{KL}\!\left(\hat{p}(\cdot\mid x_b)\,\big\|\,p_\phi\!\left(\cdot\mid \operatorname{sg}(H^{(L-2)}_\theta(x_b))\right)\right)
\]

Here $\theta$ parameterizes the convolutional layers, $\phi$ parameterizes the classification head, and $\operatorname{sg}(\cdot)$ denotes the stop-gradient operator. \emph{This operator prevents KL gradients from flowing into the convolutional layers}—only the head parameters $\phi$ receive gradients from the KL term. The convolutional representations are thus shaped entirely by the MSE regression targets, avoiding collapse to standard backpropagation. This layer-wise MSE objective bears similarity to target propagation methods \citep{lee2015difference}, although our targets come from different sources.

\paragraph{Capacity of the blocks.}
Target maps of intermediate layers produced by search are very high dimensional and involve large representational jumps. We empirically find that increased network capacity is helpful to learn these searched representations. While search uses blocks with 2 convolutional layers each, during learning we expand each block to contain 6 convolutional layers (tripling the depth). This additional capacity allows the network to better fit the complex representational targets discovered by search. To ensure fair comparison, we also test SGD baselines with 2, 4, and 6 convolutional layers per block (1$\times$, 2$\times$, and 3$\times$ the original depth) and report the best-performing configuration.

\paragraph{Supervision variants.}
By default, we supervise all three convolutional blocks with MSE losses and apply KL loss on the logits—we call this variant "Search-based (All layers)". Because earlier layers were difficult to optimize using search-based learning, we also tested an alternative approach: we skip direct supervision on convolutional block 0 and let it learn indirectly through gradients backpropagated from MSE losses in later blocks. We call this variant "Search-based (Skip block 0)". These variants are compared only in the results with data augmentation.

\subsection{Results}
To assess whether regressing to searched representations yields competitive generalization, we compare against standard stochastic gradient descent (SGD) training on MNIST, CIFAR-10, and CIFAR-100. Results are reported both with and without data augmentation to evaluate performance under minimal regularization. For MNIST, augmentation is omitted since performance already saturates without it. Since our search-based training does not use label smoothing, we also omit it from the SGD baselines for fair comparison. In the augmented setting, we cache each data point's searched representation once and reuse it across all augmented variants for efficiency.

\cref{tab:test-acc-noaug} shows that without data augmentation, our method achieves test accuracy within 1\% of SGD across all three benchmarks. With data augmentation (\cref{tab:test-acc-aug}), the variant that supervises all layers (Search-based (All layers)) performs worse than SGD, trailing by a few points on CIFAR-10 and CIFAR-100. However, the variant that skips supervision on the first block (Search-based (Skip block 0)) performs substantially better—trailing SGD by just 1.0\% on CIFAR-10 and 2.6\% on CIFAR-100. This configuration allows the network body to still benefit from the searched representations, though the remaining gap indicates further improvements are needed. These results show that search-based regression can achieve competitive generalization with standard SGD training, even though it relies on cached representations.

\begin{table}[t]
\centering
\caption{Test accuracies (\%) without data augmentation. All results are reported as mean$\;\pm\;$std over 3 independent runs.}
\label{tab:test-acc-noaug}
\vspace{0.6em}
\begin{tabular}{lccc}
\hline
 & MNIST & CIFAR-10 & CIFAR-100 \\
\hline
SGD & $99.1 \pm 0.1$ & $89.1 \pm 0.3$ & $62.3 \pm 0.5$ \\
Search-based Learning & $99.0 \pm 0.1$ & $88.3 \pm 0.3$ & $61.6 \pm 0.2$ \\
\hline
\end{tabular}
\end{table}

\begin{table}[t]
\centering
\caption{Test accuracies (\%) with data augmentation. Mean$\;\pm\;$std over 3 runs. Search-based (Skip block 0) does not apply supervision to convolutional block 0; Search-based (All layers) supervises all blocks.}
\label{tab:test-acc-aug}
\vspace{0.6em}
\begin{tabular}{lcc}
\hline
 & CIFAR-10 & CIFAR-100 \\
\hline
SGD & $93.0 \pm 0.2$ & $71.8 \pm 0.4$ \\
Search-based (Skip block 0) & $92.0 \pm 0.1$ & $69.2 \pm 0.3$ \\
Search-based (All layers) & $90.6 \pm 0.3$ & $66.5 \pm 0.3$ \\
\hline
\end{tabular}
\end{table}

\subsection{Effects of scaling search on learning}
We saw in \cref{subsec:compute-scaling} that the fitness of searched representations improves with more search compute. Now, we investigate how the validation accuracy behaves when we train on representations obtained with increased search. Again, we scale search by varying the population size or number of generations while holding the other fixed. Then we run our training procedure on the cached representations formed by our search procedure. 

\begin{figure}[t]
\centering
\begin{minipage}{0.49\linewidth}
    \includegraphics[width=\linewidth]{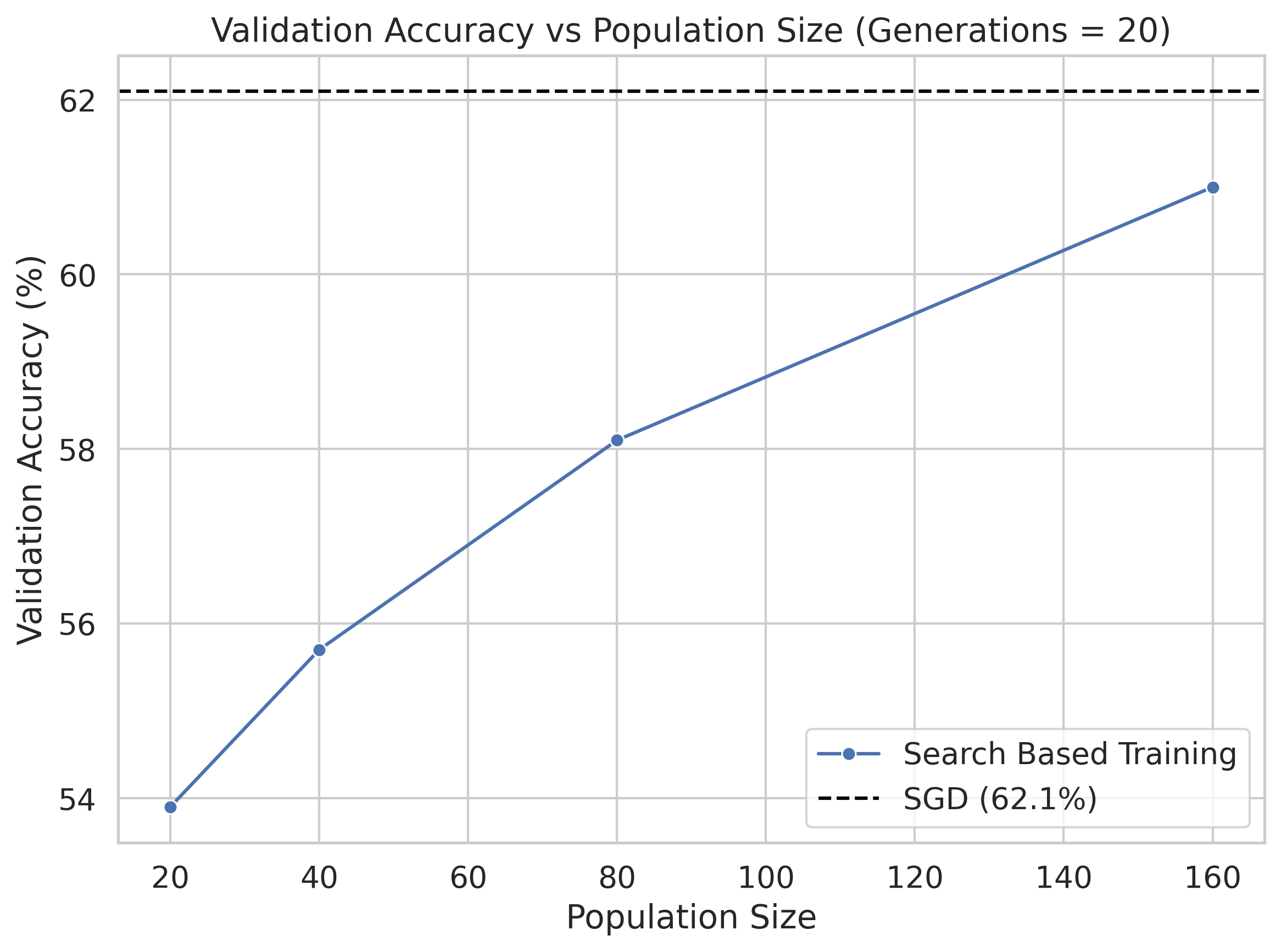}
\end{minipage}\hfill
\begin{minipage}{0.49\linewidth}
    \includegraphics[width=\linewidth]{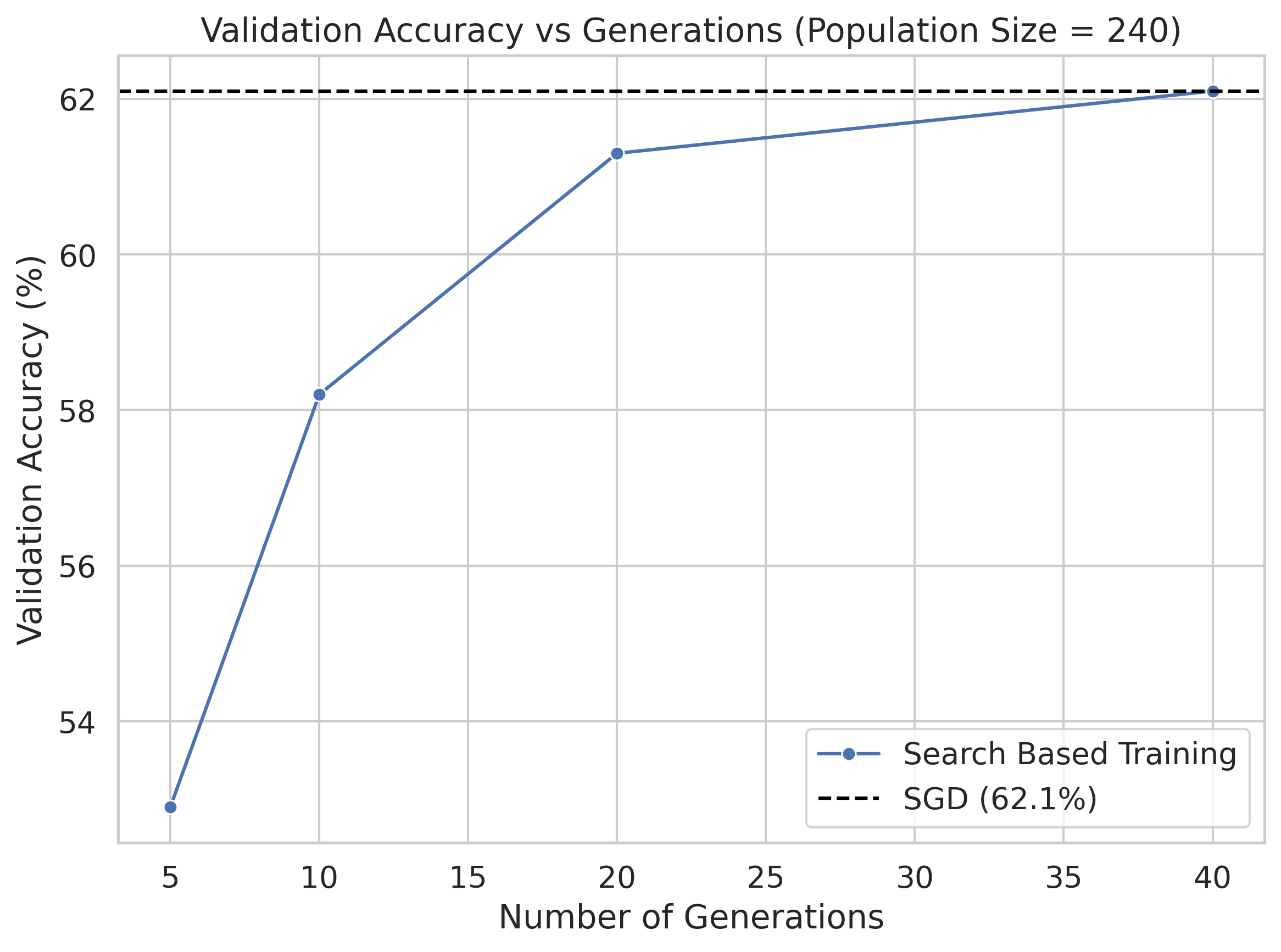}
\end{minipage}
\caption{Accuracy scales with compute usage in evolutionary search on CIFAR-100 (no data augmentation).
Validation Accuracy vs.\ population size (left) and vs.\ generations (right).}
\label{fig:learning-scaling}
\end{figure}

We report the results in \cref{fig:learning-scaling}. Observe that the accuracy scales with more compute along both the number of generations and the population size.

\subsection{Comparing models trained with Search-based learning and SGD}
\label{subsec:diff-models}
The next natural question is whether the models trained with search-based learning are different from models trained with SGD. We intuitively expect this to be the case given the drastic difference between gradient-based learning in parameter space and search-based learning. Given the difficulty of measuring distance on models in parameter space, we instead measure the distance on the representations that trained models produce. We measure cosine distance during training on the validation set between the searched target representations and representations from models trained either with search-based learning or SGD. The cosine distance is computed on the activations after flattening the feature maps into vectors: a $256 \times 15 \times 15$ feature map for block 0, a $256 \times 7 \times 7$ feature map for block 1, and a $256 \times 3 \times 3$ feature map for block 2. Cosine distance is one minus cosine similarity, ranging from 0 (identical) to 2 (opposite).

\begin{figure}[t]
    \centering
    \begin{minipage}{0.32\linewidth}
        \centering
        \includegraphics[width=\linewidth]{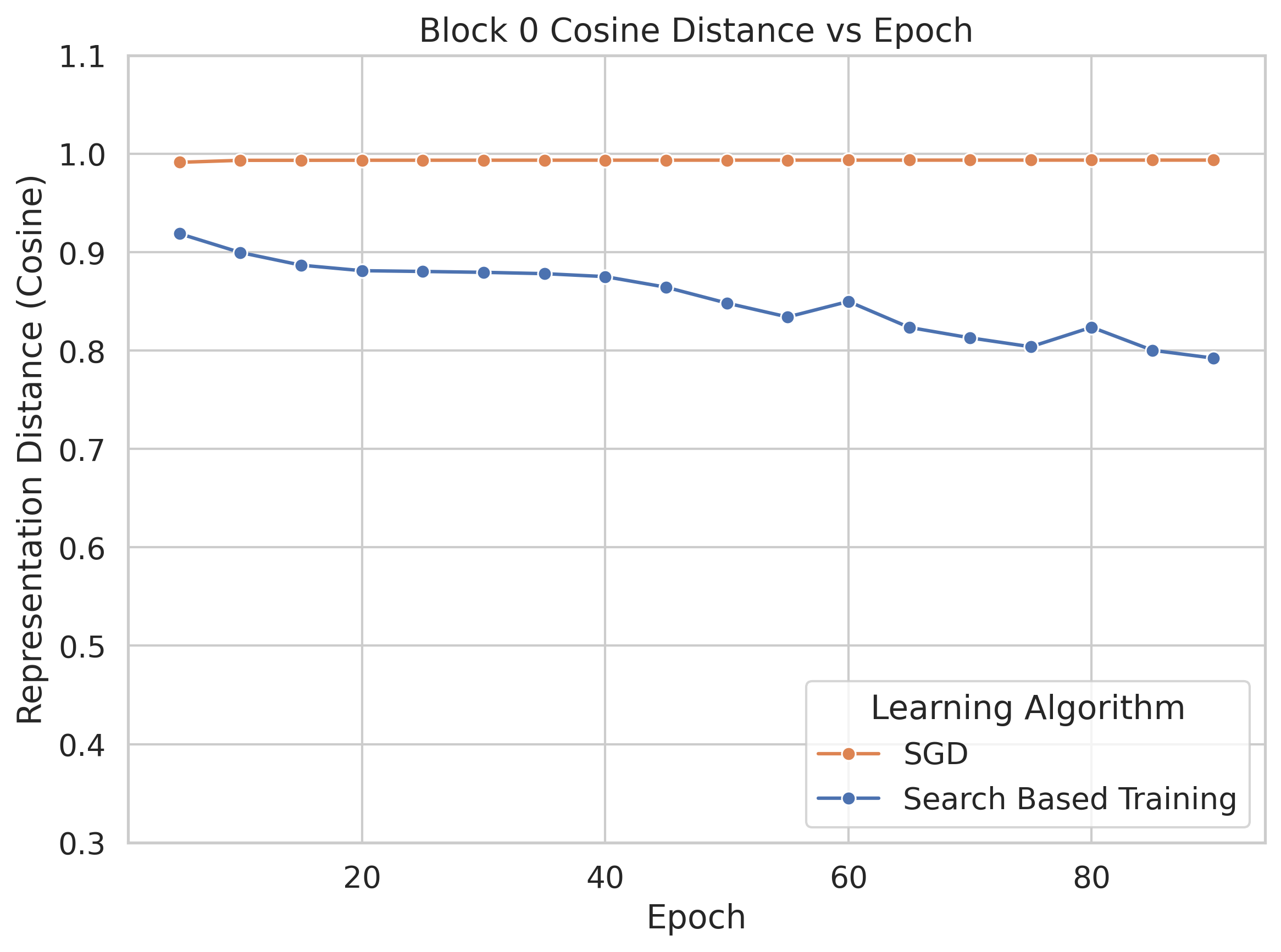}
        \\ Cosine distance at block 0.
        \label{fig:cosine-layer0}
    \end{minipage}\hfill
    \begin{minipage}{0.32\linewidth}
        \centering
        \includegraphics[width=\linewidth]{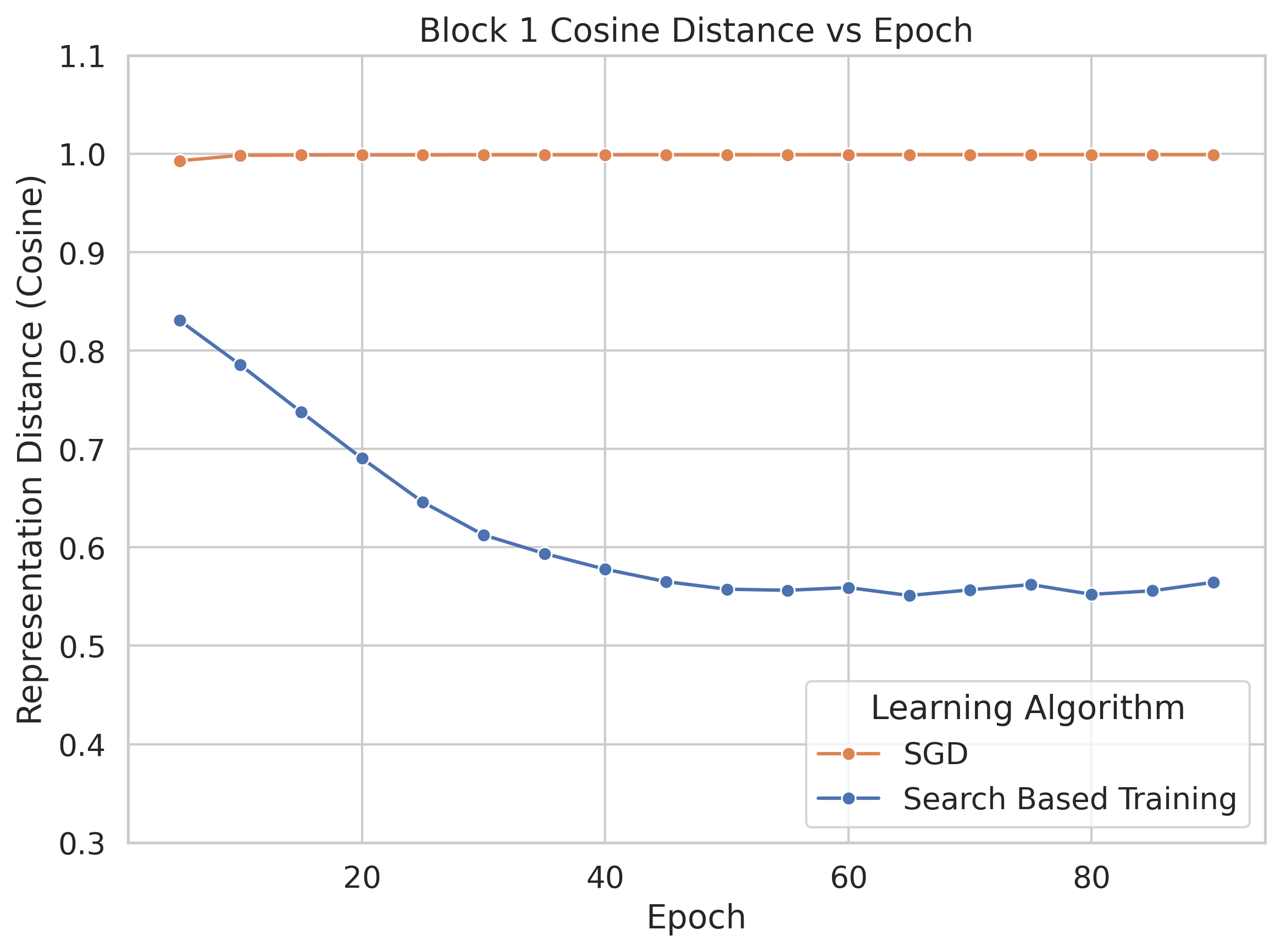}
        \\ Cosine distance at block 1.
        \label{fig:cosine-layer1}
    \end{minipage}\hfill
    \begin{minipage}{0.32\linewidth}
        \centering
        \includegraphics[width=\linewidth]{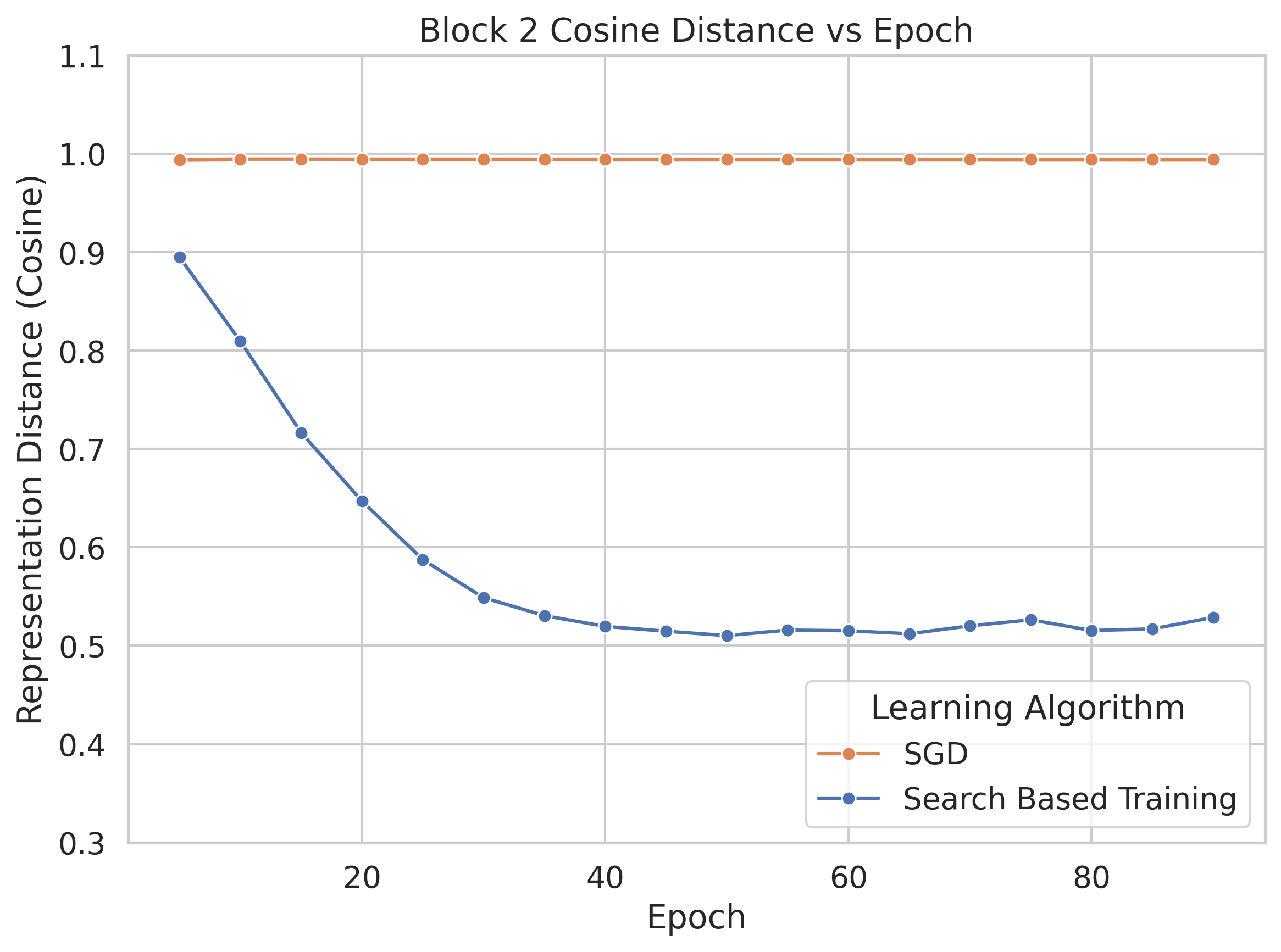}
        \\ Cosine distance at block 2.
        \label{fig:cosine-layer2}
    \end{minipage}
    \caption{Cosine distance to searched representations vs. epoch at different searched layers with search-based training and SGD on CIFAR-100 (validation set, no data augmentation).}
    \label{fig:model-distance}
\end{figure}

The results are shown in \cref{fig:model-distance}. Across all three layers, the cosine distance between SGD and the searched targets remains large—close to 1.0—indicating that SGD converges to representational solutions that are quite different from those produced by search. At the same time, the plots show that the search-based training itself is well supervised across layers, with the cosine distance lowering over the course of training.

In addition to cosine distance, we also use collision entropy 
(defined in \cref{sec:search}) to compare search-based learning and SGD. 
We calculate two types of collision entropy: within-class (averaging pairs of examples that belong to the same class) and between-class (averaging pairs of examples that belong to different classes). Lower values indicate more similar representations, while higher values indicate more distinct representations. This provides a 
complementary view of how the two training methods organize representation space.

\begin{figure}[t]
\centering
\begin{minipage}{0.48\linewidth}
    \centering
    \includegraphics[width=\linewidth]{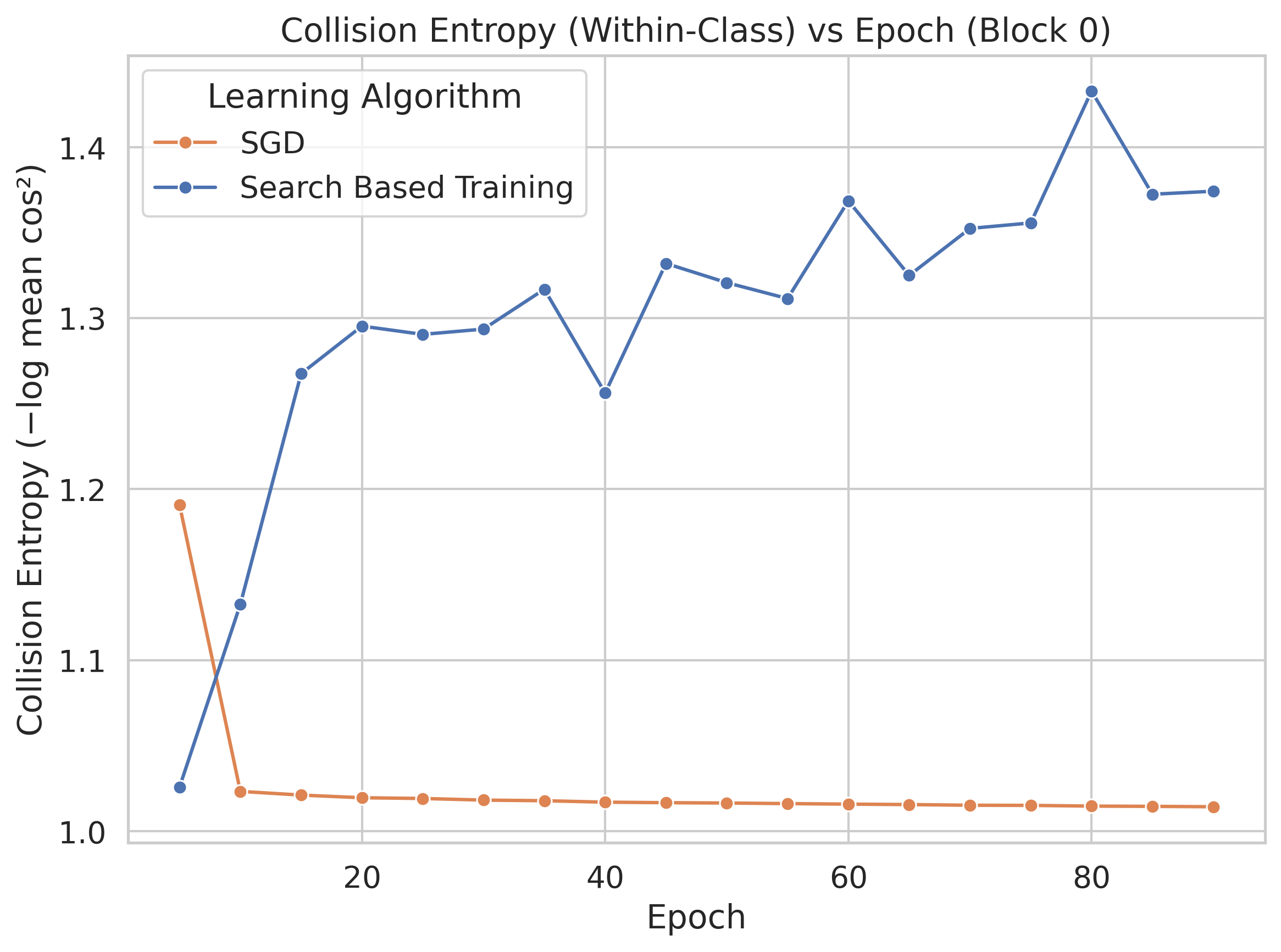}
    \\ Within-class (Block 0)
\end{minipage}\hfill
\begin{minipage}{0.48\linewidth}
    \centering
    \includegraphics[width=\linewidth]{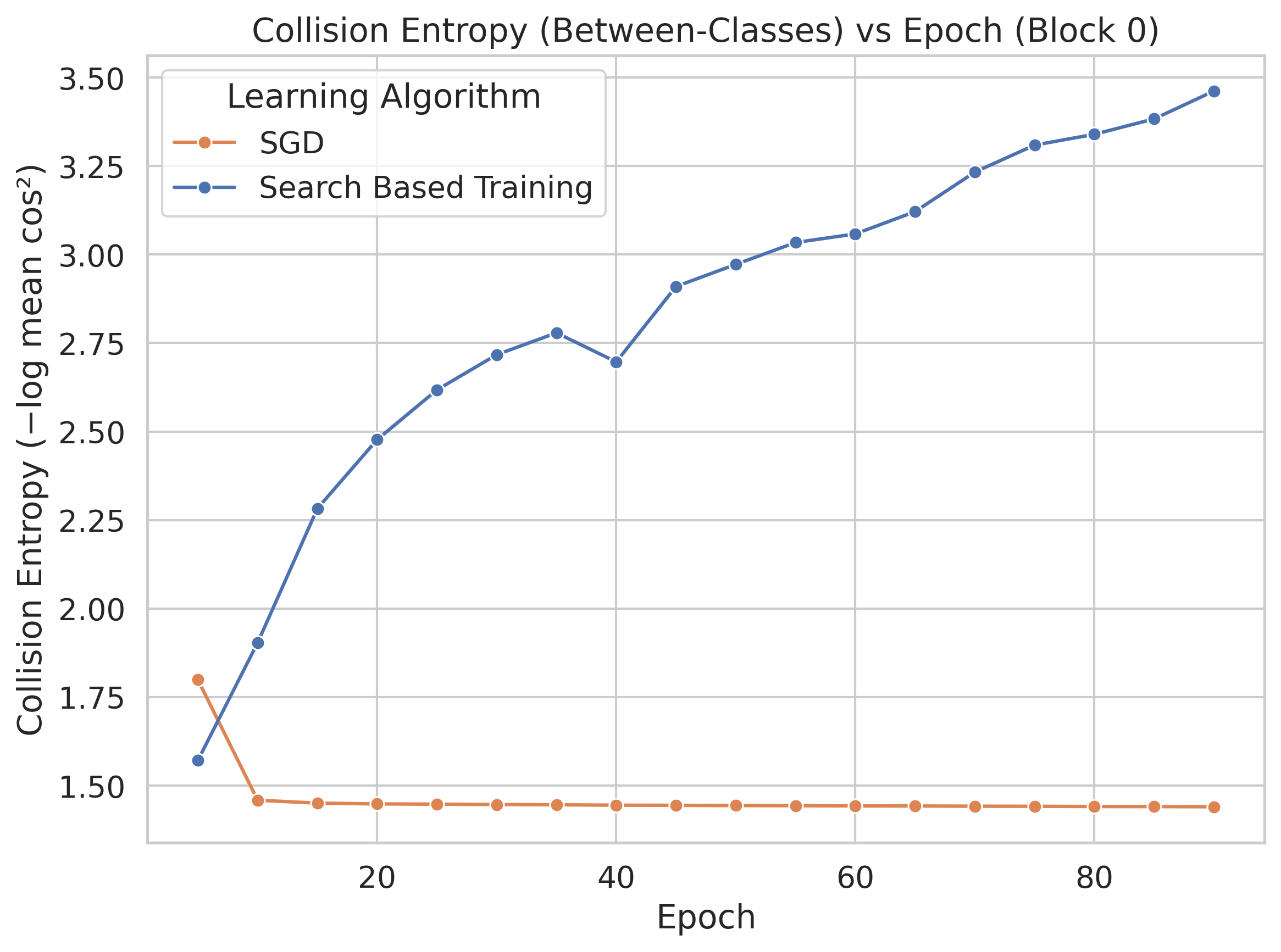}
    \\ Between-class (Block 0)
\end{minipage}
\caption{Collision entropy within and between classes for SGD and search-based training (Block 0, validation set, no data augmentation). 
Plots for Blocks 1 and 2 are deferred to Appendix~\ref{app:collision-entropy}. Experiments performed on CIFAR-100.}
\label{fig:collision-entropy-training-layer0}
\end{figure}

\cref{fig:collision-entropy-training-layer0} shows the trajectories of within-class and between-class collision entropy during training at block 0. The patterns differ noticeably between search-based training and SGD. For example, representations from search-based training tend to form a more distinct separation in block 0 between classes.
Plots for blocks 1 and 2 are included in \cref{app:collision-entropy}.
These differences highlight that the learning dynamics of the two approaches are not the same, even though both produce meaningful organization of the representation space.

\section{Related Work}
\paragraph{Alternatives to backpropagation.}
Many alternatives to backpropagation focus on improving biological plausibility. Target propagation and its variants are the most similar to our method, which similarly does regression on layerwise targets \citep{lee2015difference, bengio2014targetprop, ernoult2022scalingdtp}. Our method differs in that the targets are obtained through evolutionary search. The Forward-Forward method is an alternative to backpropagation which trains layers contrastively without any backward pass \citep{hinton2022forwardforward}. Feedback alignment is another alternative to backpropagation that uses fixed random feedback connections rather than exact gradient signals from transposed forward weights \citep{lillicrap2016feedbackalignment}. These alternatives to backpropagation change the learning mechanism for biological plausibility, unlike our method which aims to discover diverse models. 

\paragraph{Search-based training methods.}
A parallel line of research uses search in neural net training rather than relying solely on gradient descent. Many neuroevolution methods search in parameter space. For example, Natural Evolution Strategies evolve a search distribution over parameters \citep{wierstra2014nes}. NEAT takes a broader approach by searching over both parameters and network topologies \citep{stanley2002neat}. Searching in parameter space can match gradient-based methods on certain reinforcement learning tasks \citep{salimans2017es}, and this method scales to networks with millions of parameters with massive parallelization \citep{such2017deepneuroevo}. However, these methods face computational challenges due to the high dimensionality of the parameter space. Other approaches avoid this limitation by searching in latent space. Latent Program Network searches in continuous latent program space with gradient descent \citep{bonnet2024lpn}, and other work has also searched in a learned latent program space to maximize environment rewards \citep{trivedi2021learning}. Our method performs search on the hidden activations of a network directly, rather than using a learned embedding space.

\section{Conclusion, Limitations, and Future Work}

We demonstrated that neural network training can be decomposed into search over representations and gradient-based learning to match those representations. Our method achieves performance comparable to SGD while following fundamentally different optimization paths—the network body learns exclusively from searched representations rather than from classification gradients. This work provides a proof of concept that tractable search in representation space can guide parameter optimization, potentially addressing gradient descent's fundamental limitation of converging to a single solution.

Our approach has two main limitations. First, while our performance is comparable to SGD, we still trail behind it—our method is not yet a complete replacement for gradient descent. Future work must bridge this gap for the method to be viable in practice. Second, we currently use one-shot search with cached representations rather than iterative cycles where search and learning inform each other. Future work should implement tight feedback loops—networks learn searched representations, then trained networks inform the next search iteration. Additionally, while our search produces diverse representations (\cref{subsec:compute-scaling}) that lead to distinct models compared to SGD (\cref{subsec:diff-models}), more research is needed on how diverse representations translate to diverse model solutions—the ultimate goal.

\bibliographystyle{iclr2025_conference}
\bibliography{references}

\clearpage
\appendix

\section{Implementation Details}

\subsection{Model Architecture}

We adapt the optimized CIFAR architecture from \url{https://github.com/KellerJordan/cifar10-airbench} with two changes: (1) first block widened from 64 to 256 channels, (2) final max pool removed in favor of a larger linear layer.

\textbf{Overall structure:} Whitening layer $\rightarrow$ 3 convolutional blocks $\rightarrow$ linear classifier. The blocks produce feature maps of $256 \times 15 \times 15$, $256 \times 7 \times 7$, and $256 \times 3 \times 3$ respectively.

\textbf{Whitening:} $3 \rightarrow 24$ channels, kernel size 2, learned via eigendecomposition of 5000 training patches.

\textbf{ConvGroup structure:} Conv($3 \times 3$) $\rightarrow$ MaxPool($2 \times 2$) $\rightarrow$ BatchNorm $\rightarrow$ GELU $\rightarrow$ Conv($3 \times 3$) $\rightarrow$ BatchNorm $\rightarrow$ GELU.

\textbf{Convolutional blocks:} Each of the 3 convolutional blocks consists of:
\begin{itemize}[leftmargin=1.25em]
\item 1 ConvGroup with pooling (reduces spatial dimensions)
\item 2 additional ConvGroups without pooling (increases capacity for learning searched representations)
\end{itemize}
This gives each block 3 ConvGroups total, tripling the capacity compared to a single ConvGroup baseline.

\textbf{Classification head:} Linear layer mapping flattened features to class logits, with output scaling factor 1.0.

\subsection{Search Hyperparameters}

\textbf{Population structure:} 240 total candidates, selecting 20 parents per image, generating 100 exploratory children (high mutation) and 120 refinement children (standard mutation).

\textbf{Evolution schedule:}
\begin{itemize}[leftmargin=1.25em]
\item Layer 0: 300 generations, mutation strength $\alpha=0.3$, exploration boost $5.0\times$, 5 blur passes
\item Layer 1: 100 generations, $\alpha=0.2$, exploration boost $3.0\times$, 1 blur pass
\item Layer 2: 100 generations, $\alpha=0.1$, exploration boost $3.0\times$, 1 blur pass
\item Logits: 10 generations, $\alpha=0.1$, exploration boost $1.0\times$, no blur
\end{itemize}

\textbf{Selection:} Top-$k$ selection performed independently per image rather than across the batch.

\subsection{Training Hyperparameters}

\textbf{Optimization:} SGD with learning rate 4.0, momentum 0.85 with Nesterov acceleration.

\textbf{Loss weights:} MSE coefficient 1.0 for each convolutional layer, KL divergence coefficient $\lambda=0.03$ for logits (with stop-gradient preventing backpropagation through the network body).

\textbf{Batching:} Batch size 2000 with deterministic batch composition across epochs.

\textbf{Network regularization:} BatchNorm momentum 0.6, no running statistics. No label smoothing (our search-based training does not use label smoothing, so we disable it for SGD for fairness).

\textbf{Data augmentation:} Random horizontal flips and random crops from images padded by 2 pixels.

\textbf{Validation strategy:} Hold out 4000 training samples (8\% of 50k) for model selection during training, evaluating the best model on the test set only at the end.

\textbf{Training duration:} 90 epochs.

\subsection{Reproduction Operator Details}

This section provides the mathematical specifications for the genetic operators described in \cref{subsec:how-search}.

\paragraph{Crossover.}
For two randomly selected parents $H_{p_1}^{(\ell)}$ and $H_{p_2}^{(\ell)}$ at layer $\ell$:
\[
\tilde{H}^{(\ell)} = \frac{1}{2}(H_{p_1}^{(\ell)} + H_{p_2}^{(\ell)})
\]

\paragraph{Mutation.}
The mutation process differs between convolutional layers and logits.

\textbf{Convolutional layers} ($\ell < 3$): Generate channel-wise Gaussian noise $\epsilon_{b,c,h,w} \sim \mathcal{N}(0, (\alpha \sigma_{b,c})^2)$, where $\sigma_{b,c}$ is the average standard deviation of the two parents for batch element $b$ and channel $c$. Apply mutation: $\tilde{H}^{(\ell)} = \tilde{H}^{(\ell)} + \epsilon$.

\textbf{Logits layer} ($\ell = 3$): Generate element-wise Gaussian noise $\epsilon_{b,k} \sim \mathcal{N}(0, (\alpha \sigma_{b})^2)$, where $\sigma_{b}$ is the average standard deviation of the two parents across classes. Apply mutation and recenter: $\tilde{H}^{(\ell)} = \tilde{H}^{(\ell)} + \epsilon - \text{mean}(\tilde{H}^{(\ell)})$.

\paragraph{Spatial Smoothing (Convolutional layers only).}
Apply repeated $3 \times 3$ average pooling with padding to reduce high-frequency artifacts. 

\paragraph{Normalization (Convolutional layers only).}
After mutation, normalize each sample to zero mean and unit variance:
\[
\tilde{H}_{b,:,:,:}^{(\ell)} \leftarrow \frac{\tilde{H}_{b,:,:,:}^{(\ell)} - \mu_b}{\sqrt{v_b + \epsilon}}
\]
where $\mu_b$ and $v_b$ are the mean and variance computed over channels and spatial dimensions for batch element $b$, and $\epsilon = 10^{-5}$ for numerical stability.

\subsection{MNIST Hyperparameters}

We use the same search configuration as CIFAR-10 with one modification: we reduce mutation strength to $\alpha=0.1$ uniformly across all layers (layers 0–2 and logits), compared to the layer-specific values used for CIFAR. We also increase the training duration to 1000 epochs. During learning, the MNIST model uses four convolutional layers (vs.\ six for CIFAR). All other hyperparameters remain identical to the CIFAR configuration.

\section{Additional Collision Entropy Plots}
\label{app:collision-entropy}

\begin{figure}[h]
\centering

\begin{minipage}{0.48\linewidth}
    \centering
    \includegraphics[width=\linewidth]{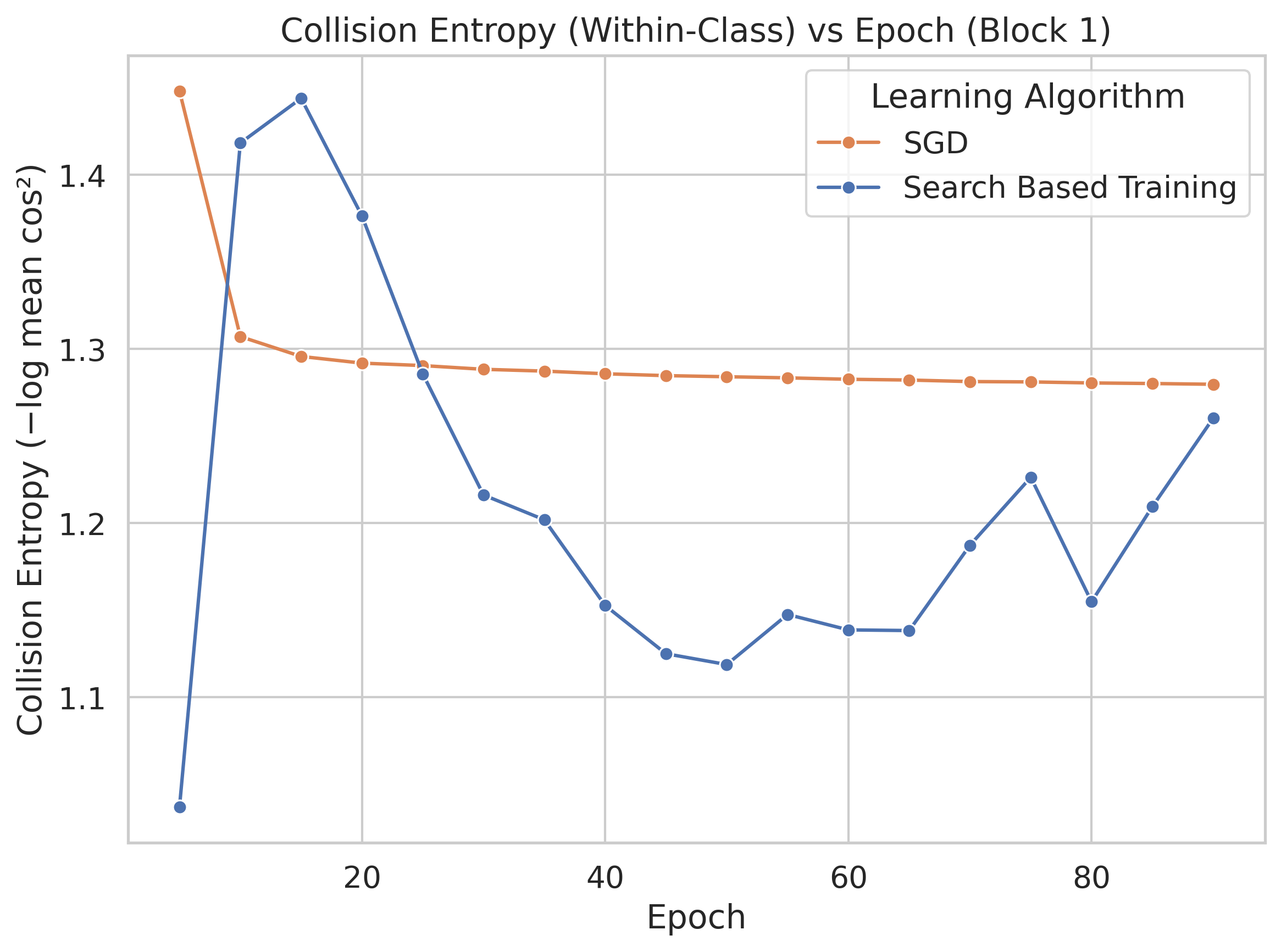}
    \\ Within-class (Block 1)
\end{minipage}\hfill
\begin{minipage}{0.48\linewidth}
    \centering
    \includegraphics[width=\linewidth]{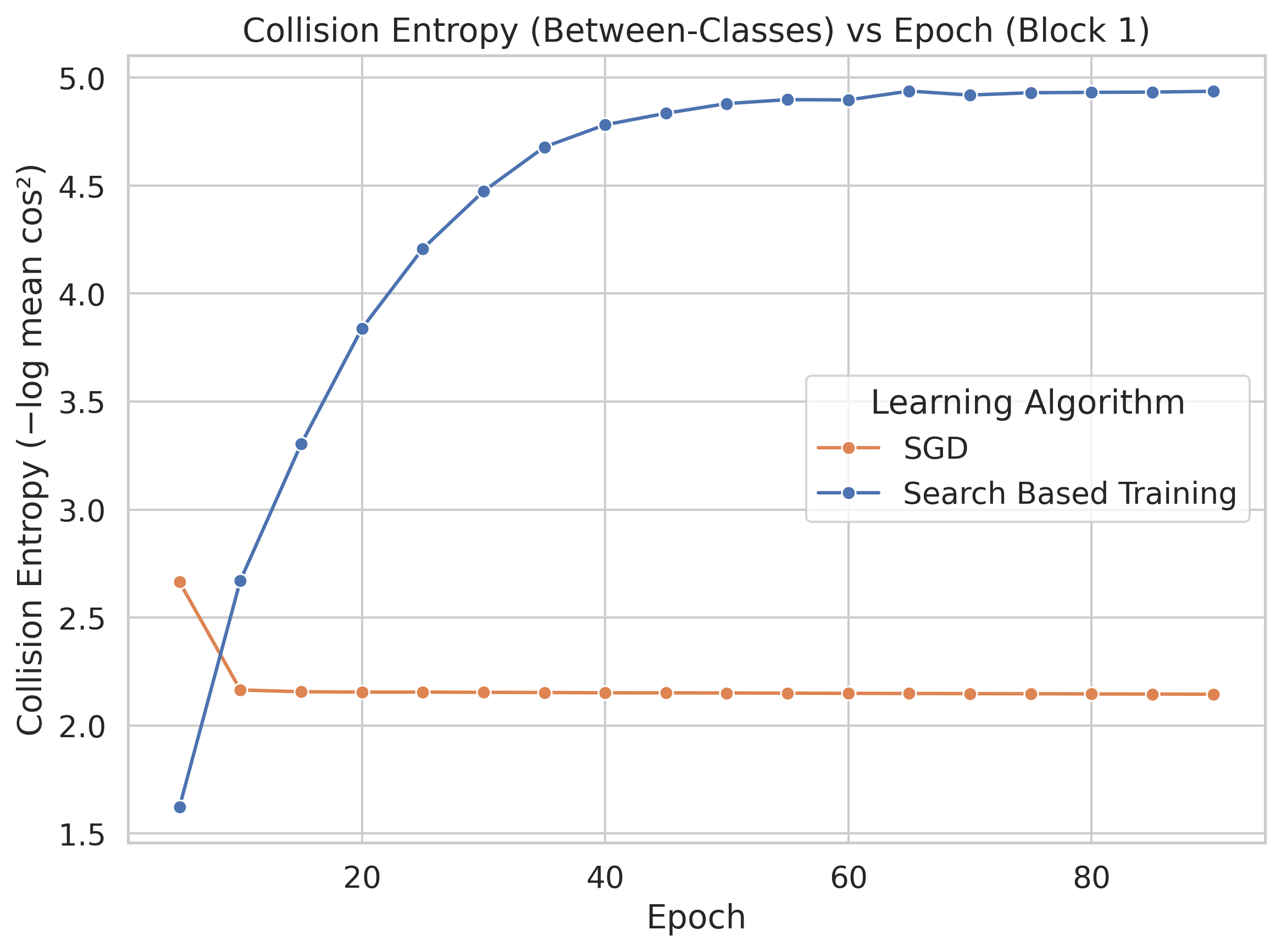}
    \\ Between-class (Block 1)
\end{minipage}

\vspace{1em}

\begin{minipage}{0.48\linewidth}
    \centering
    \includegraphics[width=\linewidth]{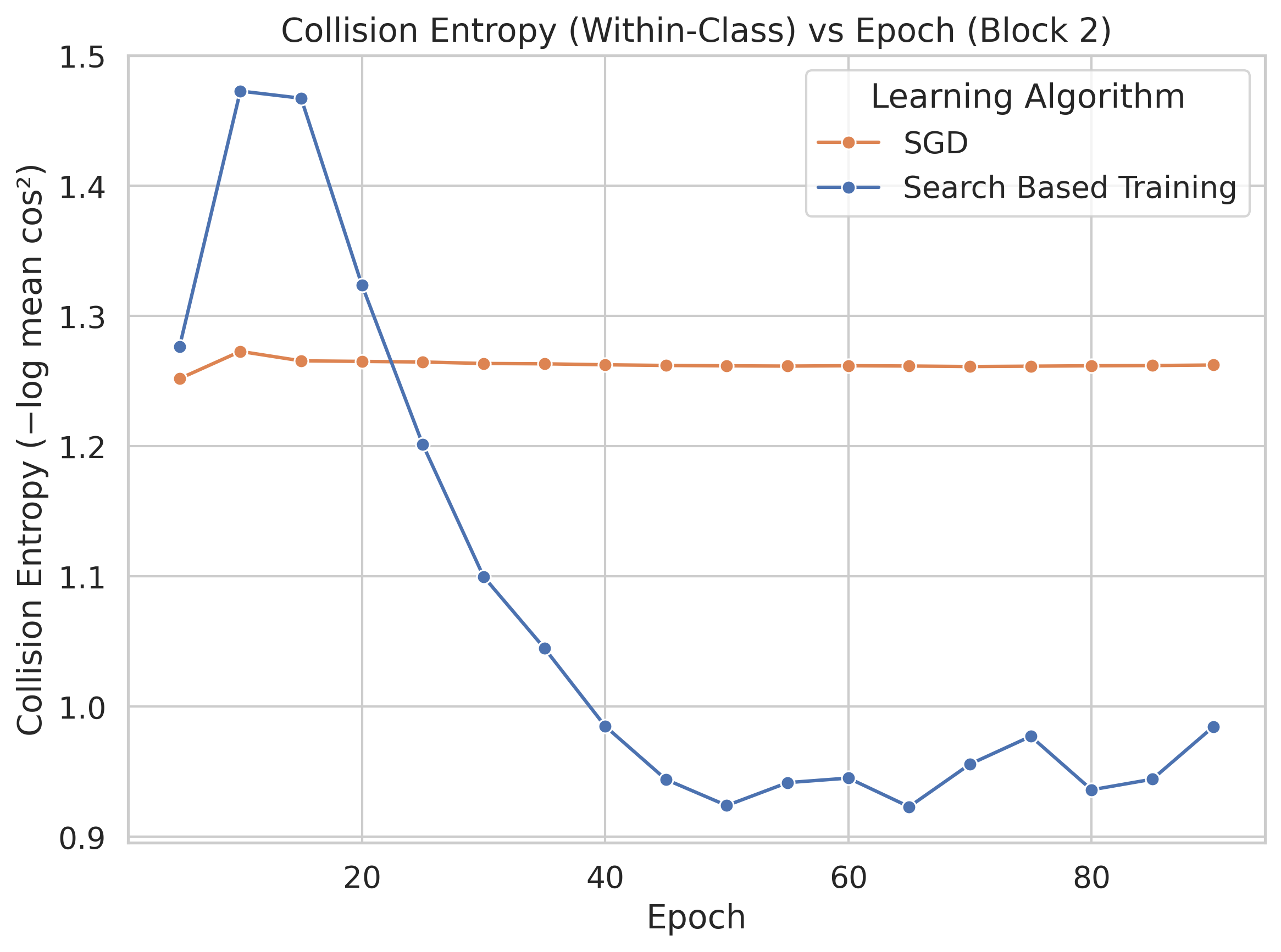}
    \\ Within-class (Block 2)
\end{minipage}\hfill
\begin{minipage}{0.48\linewidth}
    \centering
    \includegraphics[width=\linewidth]{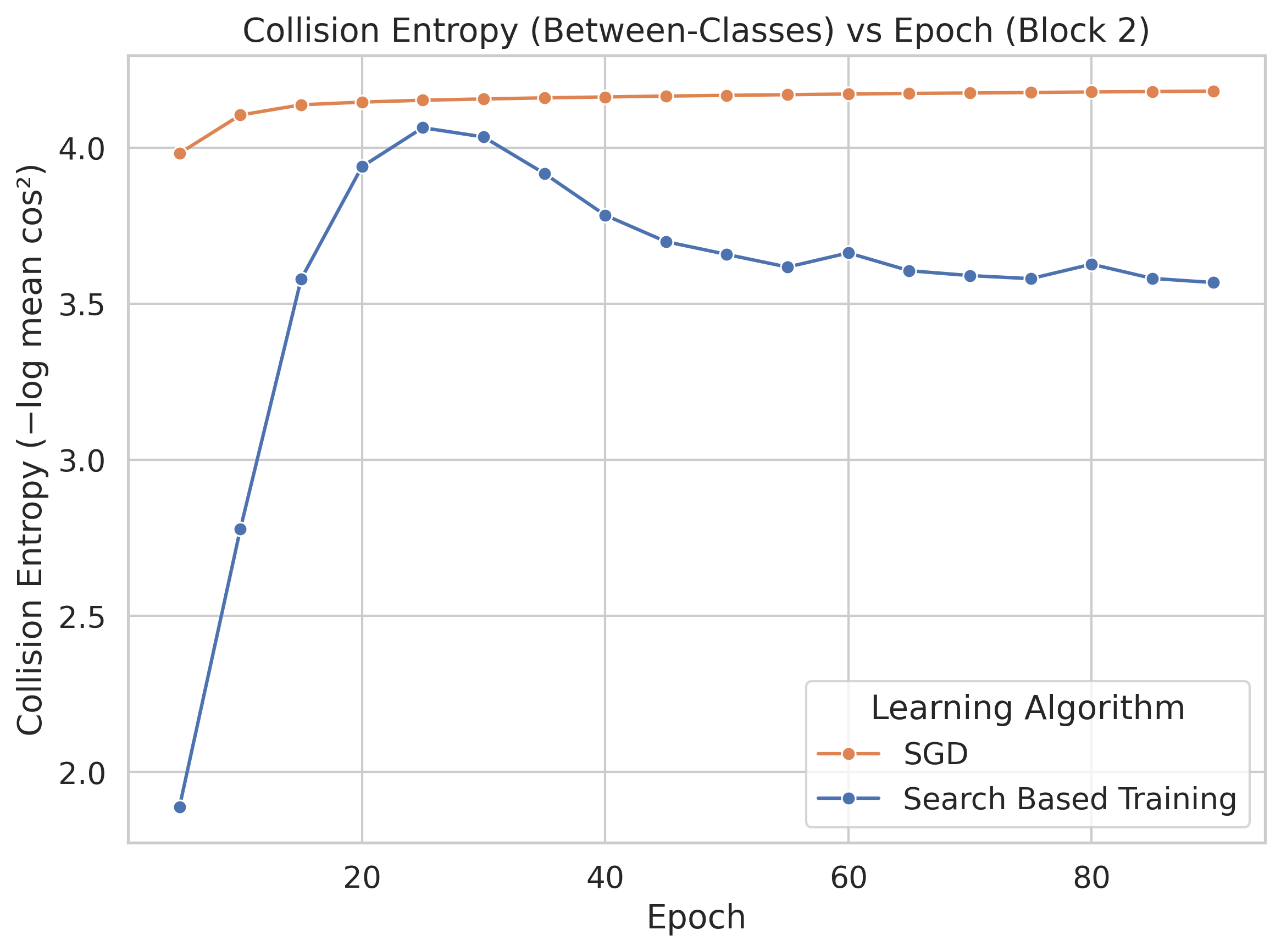}
    \\ Between-class (Block 2)
\end{minipage}

\caption{Collision entropy within and between classes for SGD and search-based training at Block 1 and Block 2 on CIFAR-100 (validation set, no data augmentation).}
\label{fig:collision-entropy-training-deeper}
\end{figure}

\end{document}